\pdfoutput=1

\documentclass[11pt]{article}

\usepackage[final]{acl}

\usepackage{times}
\usepackage{latexsym}

\usepackage[T1]{fontenc}

\usepackage[utf8]{inputenc}

\usepackage{microtype}

\usepackage{inconsolata}

\usepackage{graphicx}

\usepackage{colortbl}
\usepackage{float}
\title{ConText at WASSA 2024 Empathy and Personality Shared Task:\\ History-Dependent Embedding Utterance Representations\\ for Empathy and Emotion Prediction in Conversations}

\author{Patrícia Pereira\textsuperscript{1,2}, {\bf Helena Moniz\textsuperscript{1,3}} \and {\bf João Paulo Carvalho\textsuperscript{1,2}}\\
  \textsuperscript{1} INESC-ID, Lisbon \\
  \textsuperscript{2} Instituto Superior Técnico, University of Lisbon \\
  \textsuperscript{3} Faculdade de Letras, University of Lisbon \\
  \texttt{patriciaspereira@tecnico.ulisboa.pt} \\}

\begin{document}
\maketitle
\begin{abstract}
Empathy and emotion prediction are key components in the development of effective and empathetic agents, amongst several other applications. The WASSA shared task on empathy and emotion prediction in interactions presents an opportunity to benchmark approaches to these tasks.
Appropriately selecting and representing the historical context is crucial in the modelling of empathy and emotion in conversations. In our submissions, we model empathy, emotion polarity and emotion intensity of each utterance in a conversation by feeding the utterance to be classified together with its conversational context, i.e., a certain number of previous conversational turns, as input to an encoder Pre-trained Language Model, to which we append a regression head for prediction. We also model perceived counterparty empathy of each interlocutor by feeding all utterances from the conversation and a token identifying the interlocutor for which we are predicting the empathy.
Our system officially ranked $1^{st}$ at the CONV-turn track and $2^{nd}$ at the CONV-dialog track.
\end{abstract}

\section{Introduction}

Empathy and emotion prediction are crucial components in the development of effective and empathetic agents. There is a considerable effort to put forward modules that efficiently recognize empathy and emotion, in both users and agents from conversations and text pertaining to the most varied domains, since this knowledge can be leveraged in opinion mining, marketing, customer support, therapeutic practices, amongst other scenarios. 

The WASSA shared task on empathy and emotion prediction in interactions \cite{barriere2023findings, giorgi2024findings} presents an opportunity to benchmark approaches to these tasks.


 

In conversation pertaining tasks, knowledge of the relevant history of the conversation, i.e., the relevant previous conversational turns, is extremely useful in identifying interlocutor traits \cite{poria2019emotion, pereira2022deep}.

The usual approach to model this history has been to produce history independent representations of each utterance and subsequently perform joint modeling of those representations. State-of-the art approaches start by resorting to embedding representations from language models and employ gated, graph neural network or a combination of both architectures to perform joint modelling of these embedding representations at a later step \cite{li-etal-2021-past-present, shen-etal-2021-directed}. However, it is our contention that the Transformer, the backbone of these language models, is better at preserving the history information since it has a shorter path of information flow than the RNNs typically used for joint modelling. Following \citet{pereira-etal-2023-context}, we produce history-dependent embedding representations of each utterance, by feeding not only the utterance but also its relevant previous utterances, that pertain to the task, to the language model. We thus discard the need to deal with joint modelling after obtaining the embeddings since these constitute already an efficient representation of such history.

The results on the test set of our submissions on conversation pertaining tracks demonstrate the efficacy of our approach, both in selecting the appropriate conversational turns to be fed to the language model and in the way we feed these utterances. Our approach earned us the first place in the modelling of empathy, emotion polarity and emotion intensity of each utterance in a conversation and second place in the modelling of counterparty empathy, with a result very slightly below the top ranking submission.

\section{System Descriptions}

\subsection{Task Descriptions}

Given dyadic conversations, the tasks consist in:

\begin{itemize}
 
  \item \textbf{Track CONV-turn:} Modelling empathy, emotion polarity and emotion intensity of each utterance in a conversation. Each utterance in a conversation was annotated with these 3 traits, on a scale or real numbers from 0 to 5.
   \item \textbf{Track CONV-dialog:} Modelling perceived counterparty empathy of each interlocutor in a conversation. Each interlocutor of a dyadic conversation rated the perceived counterparty empathy, on a scale of integers from 1 to 7.

\end{itemize}


\subsection{History-Dependent Embedding Representations}

Embeddings from Pre-trained Language Models (PLMs) are the most commonly used state-of-the-art approaches in these tasks. RoBERTa \citep{liu2019roberta} is a PLM suceeding from BERT \citep{devlin-etal-2019-bert}, pre-trained to perform language modeling to learn deep contextual embeddings, i.e., vectors representing the semantics of each word or sequence of words.
DeBERTa \citep{he2020deberta} differentiates from the previous PLMs by introducing disentangled attention and an enhanced mask decoder. Longformer \citep{beltagy2020longformer} was conceived for tackling long texts, using modified attention mechanisms, acting on both local and global scale.

We now describe how we obtain embedding representations with the PLM. These processes are depicted in Figure \ref{f1}. For each track we leverage different representations:

\begin{figure}[!ht]
\begin{center}
  \includegraphics[width=\linewidth]{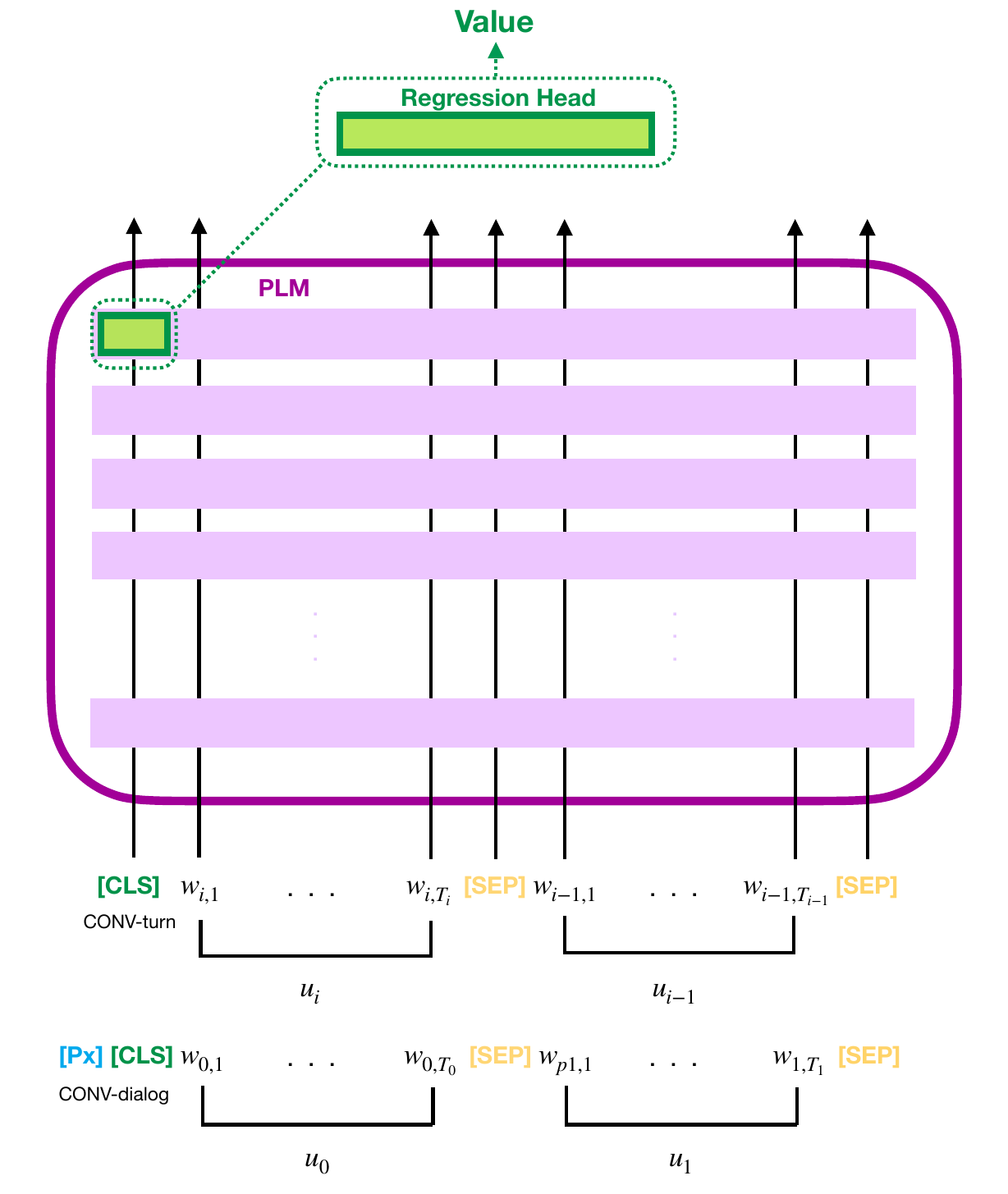}
  
 \end{center}
  \caption{Model architecture. Utterances are given as input to the PLM, of which the [CLS] token of the last layer is fed to the regression head that predicts the trait value. First input line corresponds to Track CONV-turn and second to Track CONV-dialog.
}
  \label{f1}
  \end{figure}

\begin{itemize}
  \item \textbf{Track CONV-turn:} we feed as input to the PLM the utterance we intend to classify, $u_{i}$, concatenated with its conversational context corresponding to a number $c$ of previous utterances in the conversation, $(u_{i-1},u_{i-2},...,u_{i-c})$. Concretely, we feed $u_{i}$ to the model, preceded by the [CLS] token and suceded by the [SEP] token, followed by the previous turns $u_{i-1}$ up to $u_{i-c}$, separated by the [SEP] token. An utterance consists in a sequence of $w_{it}$ tokens representing its $T_{i}$ words:

    \begin{equation}\label{equ}
    u_{i}=(w_{i1},w_{i2},...,w_{iT_{i}})
    \end{equation}

  The motivation behind feeding previous context turns lies within the fact that empathy and emotion are deeply context-dependent. Similar to human judgement in which these traits are better evaluated given the conversational context, the language model also benefits from this knowledge.

  \item \textbf{Track CONV-dialog:} we use the same backbone architecture but instead of feeding a certain number of previous utterances of the conversation, we feed all the utterances and by the order of which they were written. We also add as input in the beginning a token corresponding to the interlocutor for which we are predicting the trait.

  Since we are predicting an interlocutor trait from a dialog, it is our contention that feeding all that dialog utterances provides the language model with the most complete information and that adding a token identifying the interlocutor for which we are predicting the trait makes the model establish a distinction between the interlocutors, which is necessary since the same dialog is used twice to predict both interlocutors' trait.
  

\end{itemize}

From the obtained embeddings we can extract a suitable representation for the sentence. Choosing all tokens from all layers would yield an extremely memory demanding classification layer and may not yield the best model performance. Thus we choose the first embedding from the last layer L, the [CLS], as in Equation \ref{ecls}: 
\begin{equation}\label{ecls}
pooled_{i}=PLM_{L,[CLS]}(input_{i})
\end{equation}

The regression module that follows the PLM is a linear fully connected layer, applying a linear transformation to the pooled encoder output data, changing the dimension of this data from the PLM hidden size to 1:

\begin{equation}\label{eq6}
value_{i}=pooled_{i}W^{T}+b
\end{equation}

We then minimize the Mean Squared Error (MSE) loss between the predicted value and the gold label:
\begin{equation}\label{eq7}
loss=MSE(value_{i}, goldlabel_{i})
\end{equation}

\section{Experimental Setup}

\subsection{Training Details}

The models used are RoBERTa, DeBERTa and Longformer, all in both base and large versions from the Transformers library by Hugging Face \cite{wolf-etal-2020-transformers}. The Adam \cite{kingma2014adam} optimizer is used with an initial learning rate of 1e-5 and 5e-5, for the encoder and the regression head, respectively with a layer-wise decay rate of 0.95 after each training epoch for the encoder. The encoder is frozen for the first epoch. The batch size is set to 4. Gradient clipping is set to 1.0. As stopping criteria, early stopping is used to terminate training if there is no decrease after 5 consecutive epochs on the validation set over MSE, for a maximum of 40 epochs.  The checkpoint used for obtaining the results on the test set is the one that achieves the lowest MSE on the validation set. When running training to determine which backbone model for each trait or number of context-turns on the CONV-turn track for each trait should be used for training the final model, we use the provided validation set as test set and perform a 90:10 split of the provided training set into training and validation sets.

\vspace{2.5mm}

Our code is publicly available\footnote{\url{https://github.com/patricia-pereira/wassa-sharedtask}}.

\subsection{Dataset}

The shared task dataset consists in empathic reactions to news stories and associated conversations, containing dyadic conversations in reaction to news articles where there is harm to a person, group, or other \cite{omitaomu2022empathic}. These conversations are turn-level annotated in perceived empathy, emotion polarity, and emotion intensity, and dialogue-level annotated in terms of perceived counterparty empathy.

Given a conversation, the data is processed and fed to the model to train as depicted in Table \ref{tab:dialogue}:

\begin{table}[H]
\centering
\small
\begin{tabular}{l}




\cellcolor[HTML]{F4E9C9}\textbf{Conversation:}\\

\cellcolor[HTML]{F4E9C9}P1: its a shame with the drought \\

\cellcolor[HTML]{F4E9C9}P2: It's terrible what is happening to the world today!\\

\cellcolor[HTML]{F4E9C9}P1: I know so much distruction\\

\cellcolor[HTML]{F4E9C9}P2: Do you think it is human caused?\\

\textbf{Emotion Intensity:} 1.3333\\

\textbf{Emotion Polarity:} 1\\

\textbf{Empathy:} 1\\

\cellcolor[HTML]{F4E9C9}P1: maybe probably thoug\\

\cellcolor[HTML]{F4E9C9}P2: I wonder what will be done to fix the destruction.\\

\cellcolor[HTML]{F4E9C9}P1: probably nothing humans don't really care\\

\textbf{Perceived Empathy of P1 rated by P2:} 1\\

 \cellcolor[HTML]{F4E9C9}\textbf{Track CONV-turn (Emotion Intensity)}\\ 
 
 \cellcolor[HTML]{F4E9C9}\textbf{Input:} [CLS] Do you think it is human caused? [SEP]  \\ 

\cellcolor[HTML]{F4E9C9}I know so much  distruction [SEP]\\ 

\cellcolor[HTML]{F4E9C9}It's terrible  what is happening to the world today! [SEP]  \\ 

\textbf{Output:} 1.3333\\

\cellcolor[HTML]{F4E9C9}\textbf{Track CONV-dialog (Person 1)}\\ 

\cellcolor[HTML]{F4E9C9}\textbf{Input:} [P1][CLS] its a shame with the drought [SEP] \\

\cellcolor[HTML]{F4E9C9}It's terrible  what is happening to the world today! [SEP]  \\

\cellcolor[HTML]{F4E9C9}I know so much  distruction [SEP]\\

\cellcolor[HTML]{F4E9C9}Do you think it is human caused? [SEP]\\

\cellcolor[HTML]{F4E9C9}maybe probably thoug [SEP]\\

\cellcolor[HTML]{F4E9C9}I wonder what will be done to fix the destruction. [SEP]\\

\cellcolor[HTML]{F4E9C9}probably nothing humans don't really care [SEP]\\

\textbf{Output:} 1\\



\end{tabular}
\caption{Example of raw data and how it is given as input/output pairs to train the model}
\label{tab:dialogue}
\end{table}

\section{Results and Analysis}

\subsection{Track CONV-turn}

We now report the results of our approach for Track CONV-turn on the validation set. For representative purposes we only report the backbone model that yielded the best results, RoBERTa-large.

\begin{table}[H]
 \centering
 
  \begin{tabular}{cccc}
    \hline
       \textbf{$c$} &Polarity &Intensity &Empathy\\
      
    \hline
    $0$&0.7292&0.6242&0.6262\\
    $1$ &0.7812&0.6490&0.6688\\
    $2$ &0.7869&\textbf{0.6700}&0.6828\\
    $3$ &0.7841&0.6615&0.6815\\
    $4$&0.7828&0.6627&\textbf{0.6895}\\
   $5$&0.7912&0.6672&0.6774 \\
   $6$&\textbf{0.7928}&0.6586 &0.6727\\

    \hline
  \end{tabular} 
  \caption{Submission results for track CONV-turn on the validation set. Evolution of the Pearson correlation score with the number of appended context turns.}
 \label{trav}
\end{table} 

The lowest performance in all traits is the one obtained without introducing any context turns, highlighting the importance of considering context. The general tendency is for the performance to increase with the progressive increase of the number of context turns, up to a performance peak which is trait specific, and then for it to decrease. For the purpose of the shared task only one random seed was used to generate results, but we use 5 random seeds in our previous work \cite{pereira-etal-2023-context} to validate this tendency. The peak of performance is obtained with 6 context turns for Emotion Polarity, 2 turns for Emotion Intensity and 4 turns for Empathy.

We now report the results of our approach for Track CONV-turn on the test set and compare it with the results of other teams. 

\begin{table}[H]
 \centering
 
  \begin{tabular}{ccccc}
    \hline
       Team &Avg &Polarity &Intensity &Empathy\\
      
    \hline
    Ours& 0.626& 0.679& 0.622& 0.577\\
    $2^{nd}$&0.623 &0.680 &0.607 &0.582 \\
    $3^{rd}$&0.610 &0.671 &0.601 & 0.559\\
    $4^{th}$&0.595 &0.663 &0.589 & 0.534\\
    $5^{th}$& 0.590& 0.644&0.581 & 0.544\\
    $6^{th}$&0.588 &0.638 & 0.584&0.541\\
    $7^{th}$&0.477 & 0.422&0.473 & 0.534\\
    $8^{th}$&-0.007 &-0.018 & 0.032& 0.034\\
    $9^{th}$&-0.030 & -0.020& -0.043& -0.027\\

    \hline
  \end{tabular} 
  \caption{Submission results for track CONV-turn on the test set. Our model uses RoBERTa-large with the number of context turns which yielded the best results for each trait on the validation set.}
 \label{trat}
\end{table} 

Our team ranked first place amongst nine teams, with an average Pearson score of 0.626. Some teams scored Pearson scores very close to ours while some teams scored Pearson scores much lower and even negative. This may indicate a very diverse set of approaches resulting in different scores.

\subsection{Track CONV-dialog}

 
      


Regarding results for the track CONV-dialog on the validation set, the backbone model which yielded the best result, a Pearson correlation score of 0.3416, was RoBERTa-base. This model has a max token length of 512 and we truncate the input to respect that limit. As we feed all the conversation to the model, which usually exceeds 512 tokens, it could be expected that a backbone model such as the Longformer which has a max token length of 4016 would yield better results. These results could indicate that it is not necessary to feed all the conversation to evaluate perceived counterparty empathy. 

We now report the results of our approach for Track CONV-dialog on the test set and compare it with the results of other teams. 

\begin{table}[H]
 \centering
 
  \begin{tabular}{lc}
    \hline
     Team&Empathy\\
      
    \hline
    $1^{st}$&0.193\\
    Ours& 0.191\\
    $3^{rd}$&0.172 \\
    $4^{th}$&0.012  \\

    \hline
  \end{tabular} 
  \caption{Submission results for track CONV-dialog on the test set. Our model uses RoBERTa-base.}
 \label{trbt}
\end{table} 

We achieved a Pearson score of 0.191, just 0.002 below the top ranking submission, placing our team second in the ranking.

The result on the test set was notably lower than the result on the validation set. This can be due to the different distributions of the sets but also due to the fact that with this small dataset, the provided validation set and the validation set for choosing the model when performing the 90:10 split on the provided training set are not large enough to be representative.

\section{Discussion}

When comparing results of both tracks we observe that the result on the CONV-dialog track is significantly lower than the result on the CONV-turn track. This might be due to the fact that there are more mature approaches for emotion and empathy prediction in conversational turns, especially pertaining to the field of Emotion Recognition in Conversations \cite{pereira2022deep}, while there are less approaches and datasets for the task of predicting perceived counterparty empathy from entire conversations.




\section{Conclusion and Future Work}

We presented an efficient approach for representing the selected historical conversational context for modelling of empathy and emotion in conversations. It consisted in feeding the appropriate conversational turns as input to a PLM and resorting to a simple regression head, contrasting with approaches that feed each turn to a PLM and then perform joint modelling of the turns with more complex modules. We modelled empathy, emotion polarity and emotion intensity of each utterance in a conversation by feeding the utterance to be classified together with its conversational context and modelled perceived counterparty empathy of each interlocutor by feeding all utterances from the conversation and a token identifying the interlocutor for which we were predicting the empathy. The official results of our submissions demonstrate the efficacy of our approach, both in selecting the appropriate conversational turns to be fed to the language model and in the way we feed these utterances.

Concerning future work directions, for the task of perceived counterparty empathy, since best results were obtained with RoBERTa that only takes 512 tokens, it would be interesting to explore feeding the final 512 tokens of the conversation instead of the initial, or a different window of tokens.

\section{Limitations}

While our approach to modelling perceived counterparty empathy seems very promising when validated with the shared task dataset, given our position in the leaderboard, it still attains a modest Pearson correlation score. Furthermore, confronting with other approaches on other datasets is necessary to claim its generalization ability and suggested superiority.






\section*{Acknowledgments}
This work was supported by Fundação para a Ciência e a Tecnologia (FCT), through Portuguese national funds, Ref. UIDB/50021/2020, DOI: 10.54499/UIDB/50021/2020 and Ref. UI/BD/154561/2022 and the Portuguese Recovery and Resilience Plan, through project C645008882-00000055 (Responsible.AI).

\bibliography{anthology,custom}

\begin{thebibliography}{14}
\providecommand{\natexlab}[1]{#1}

\bibitem[{Barriere et~al.(2023)Barriere, Sedoc, Tafreshi, and Giorgi}]{barriere2023findings}
Valentin Barriere, Jo{\~a}o Sedoc, Shabnam Tafreshi, and Salvatore Giorgi. 2023.
\newblock Findings of wassa 2023 shared task on empathy, emotion and personality detection in conversation and reactions to news articles.
\newblock In \emph{Proceedings of the 13th Workshop on Computational Approaches to Subjectivity, Sentiment, \& Social Media Analysis}, pages 511--525.

\bibitem[{Beltagy et~al.(2020)Beltagy, Peters, and Cohan}]{beltagy2020longformer}
Iz~Beltagy, Matthew~E Peters, and Arman Cohan. 2020.
\newblock Longformer: The long-document transformer.
\newblock \emph{arXiv preprint arXiv:2004.05150}.

\bibitem[{Devlin et~al.(2019)Devlin, Chang, Lee, and Toutanova}]{devlin-etal-2019-bert}
Jacob Devlin, Ming-Wei Chang, Kenton Lee, and Kristina Toutanova. 2019.
\newblock \href {https://doi.org/10.18653/v1/N19-1423} {{BERT}: Pre-training of deep bidirectional transformers for language understanding}.
\newblock In \emph{Proceedings of the 2019 Conference of the North {A}merican Chapter of the Association for Computational Linguistics: Human Language Technologies, Volume 1 (Long and Short Papers)}, pages 4171--4186, Minneapolis, Minnesota. Association for Computational Linguistics.

\bibitem[{Giorgi et~al.(2024)Giorgi, Sedoc, Barriere, and Tafreshi}]{giorgi2024findings}
Salvatore Giorgi, Jo{\~a}o Sedoc, Valentin Barriere, and Shabnam Tafreshi. 2024.
\newblock Findings of wassa 2024 shared task on empathy and personality detection in interactions.
\newblock In \emph{Proceedings of the 14th Workshop on Computational Approaches to Subjectivity, Sentiment, \& Social Media Analysis}.

\bibitem[{He et~al.(2020)He, Liu, Gao, and Chen}]{he2020deberta}
Pengcheng He, Xiaodong Liu, Jianfeng Gao, and Weizhu Chen. 2020.
\newblock Deberta: Decoding-enhanced bert with disentangled attention.
\newblock \emph{arXiv preprint arXiv:2006.03654}.

\bibitem[{Kingma and Ba(2014)}]{kingma2014adam}
Diederik~P Kingma and Jimmy Ba. 2014.
\newblock Adam: A method for stochastic optimization.
\newblock \emph{arXiv preprint arXiv:1412.6980}.

\bibitem[{Li et~al.(2021)Li, Lin, Fu, and Wang}]{li-etal-2021-past-present}
Jiangnan Li, Zheng Lin, Peng Fu, and Weiping Wang. 2021.
\newblock \href {https://doi.org/10.18653/v1/2021.findings-emnlp.104} {Past, present, and future: Conversational emotion recognition through structural modeling of psychological knowledge}.
\newblock In \emph{Findings of the Association for Computational Linguistics: EMNLP 2021}, pages 1204--1214, Punta Cana, Dominican Republic. Association for Computational Linguistics.

\bibitem[{Liu et~al.(2019)Liu, Ott, Goyal, Du, Joshi, Chen, Levy, Lewis, Zettlemoyer, and Stoyanov}]{liu2019roberta}
Yinhan Liu, Myle Ott, Naman Goyal, Jingfei Du, Mandar Joshi, Danqi Chen, Omer Levy, Mike Lewis, Luke Zettlemoyer, and Veselin Stoyanov. 2019.
\newblock Roberta: A robustly optimized bert pretraining approach.
\newblock \emph{arXiv preprint arXiv:1907.11692}.

\bibitem[{Omitaomu et~al.(2022)Omitaomu, Tafreshi, Liu, Buechel, Callison-Burch, Eichstaedt, Ungar, and Sedoc}]{omitaomu2022empathic}
Damilola Omitaomu, Shabnam Tafreshi, Tingting Liu, Sven Buechel, Chris Callison-Burch, Johannes Eichstaedt, Lyle Ungar, and Jo{\~a}o Sedoc. 2022.
\newblock Empathic conversations: A multi-level dataset of contextualized conversations.
\newblock \emph{arXiv preprint arXiv:2205.12698}.

\bibitem[{Pereira et~al.(2022)Pereira, Moniz, and Carvalho}]{pereira2022deep}
Patr{\'\i}cia Pereira, Helena Moniz, and Joao~Paulo Carvalho. 2022.
\newblock Deep emotion recognition in textual conversations: A survey.
\newblock \emph{arXiv preprint arXiv:2211.09172}.

\bibitem[{Pereira et~al.(2023)Pereira, Moniz, Dias, and Carvalho}]{pereira-etal-2023-context}
Patr{\'\i}cia Pereira, Helena Moniz, Isabel Dias, and Joao~Paulo Carvalho. 2023.
\newblock \href {https://doi.org/10.18653/v1/2023.wassa-1.21} {Context-dependent embedding utterance representations for emotion recognition in conversations}.
\newblock In \emph{Proceedings of the 13th Workshop on Computational Approaches to Subjectivity, Sentiment, {\&} Social Media Analysis}, pages 228--236, Toronto, Canada. Association for Computational Linguistics.

\bibitem[{Poria et~al.(2019)Poria, Majumder, Mihalcea, and Hovy}]{poria2019emotion}
Soujanya Poria, Navonil Majumder, Rada Mihalcea, and Eduard Hovy. 2019.
\newblock Emotion recognition in conversation: Research challenges, datasets, and recent advances.
\newblock \emph{IEEE Access}, 7:100943--100953.

\bibitem[{Shen et~al.(2021)Shen, Wu, Yang, and Quan}]{shen-etal-2021-directed}
Weizhou Shen, Siyue Wu, Yunyi Yang, and Xiaojun Quan. 2021.
\newblock \href {https://doi.org/10.18653/v1/2021.acl-long.123} {Directed acyclic graph network for conversational emotion recognition}.
\newblock In \emph{Proceedings of the 59th Annual Meeting of the Association for Computational Linguistics and the 11th International Joint Conference on Natural Language Processing (Volume 1: Long Papers)}, pages 1551--1560, Online. Association for Computational Linguistics.

\bibitem[{Wolf et~al.(2020)Wolf, Debut, Sanh, Chaumond, Delangue, Moi, Cistac, Rault, Louf, Funtowicz, Davison, Shleifer, von Platen, Ma, Jernite, Plu, Xu, Le~Scao, Gugger, Drame, Lhoest, and Rush}]{wolf-etal-2020-transformers}
Thomas Wolf, Lysandre Debut, Victor Sanh, Julien Chaumond, Clement Delangue, Anthony Moi, Pierric Cistac, Tim Rault, Remi Louf, Morgan Funtowicz, Joe Davison, Sam Shleifer, Patrick von Platen, Clara Ma, Yacine Jernite, Julien Plu, Canwen Xu, Teven Le~Scao, Sylvain Gugger, Mariama Drame, Quentin Lhoest, and Alexander Rush. 2020.
\newblock \href {https://doi.org/10.18653/v1/2020.emnlp-demos.6} {Transformers: State-of-the-art natural language processing}.
\newblock In \emph{Proceedings of the 2020 Conference on Empirical Methods in Natural Language Processing: System Demonstrations}, pages 38--45, Online. Association for Computational Linguistics.

\end{thebibliography}

\end{document}